\documentclass{article}

\usepackage{arxiv}

\usepackage{microtype}
\usepackage{graphicx}
\usepackage{subfigure}
\usepackage{booktabs}

\usepackage{algorithm}
\usepackage{algorithmic}
\usepackage{times}
\usepackage{soul}
\usepackage{url}
\usepackage[utf8]{inputenc}
\usepackage{amsmath}
\usepackage{amsthm}
\usepackage{amsfonts}

\usepackage{graphicx,subfigure,caption}
\usepackage{caption}
\usepackage{wrapfig}
\usepackage{comment}
\usepackage{enumitem}
\usepackage{xcolor}
\usepackage{wrapfig}

\usepackage[natbibapa]{apacite}
\usepackage{hyperref}
\usepackage{xr}

\title{Predictive Coding: Towards a Future of Deep Learning beyond Backpropagation?}

\author{%
    Beren Millidge$^{*,1}$, Tommaso Salvatori$^{*,2}$ , Yuhang Song$^{1,2,\dagger}$, {Rafal Bogacz}$^{1}$, {Thomas Lukasiewicz}$^{2}$\\
    $^1$MRC Brain Network Dynamics Unit, University of Oxford, UK\\
    $^2$Department of Computer Science, University of Oxford, UK\\
$^1$firstname.lastname@ndcn.ox.ac.uk,
$^2$firstname.lastname@cs.ox.ac.uk
}
\begin{document}
\maketitle
\renewcommand{\thefootnote}{\fnsymbol{footnote}}
\footnotetext[1]{Equal contribution, listed in alphabetical order.}
\footnotetext[2]{Corresponding author}

\begin{abstract}

The backpropagation of error algorithm used to train deep neural networks has been fundamental to the successes of deep learning. However, it requires sequential backward updates and non-local computations, which make it challenging to parallelize at scale and is unlike how learning works in the brain. Neuroscience-inspired learning algorithms, however, such as \emph{predictive coding}, which utilize local learning, have the potential to overcome these limitations and advance beyond current deep learning technologies. While predictive coding originated in theoretical neuroscience as a model of information processing in the cortex, recent work has developed the idea into a general-purpose algorithm able to train neural networks using only local computations. In this survey, we review works that have contributed to this perspective and demonstrate the close theoretical connections between predictive coding and backpropagation, as well as works that highlight the multiple advantages of using predictive coding models over backpropagation-trained neural networks. Specifically, we show the substantially greater flexibility of predictive coding networks against equivalent deep neural networks, which can function as classifiers, generators, and associative memories simultaneously, and can be defined on arbitrary graph topologies. Finally, we  review direct benchmarks of predictive coding networks on machine learning classification tasks, as well as its close connections to control theory and applications in robotics.

\end{abstract}

\section{{Introduction}}
\label{sec:intro}

%During the deep learning revolution, started in 2012 when convolutional networks where used to reach state-of-the-art on ImageNet \citep{Krizhevsky2012}, machine learning research on neuroscience-inspired learning methods was not very active. 
Classical backpropagation (BP) 
\citep{rumelhart1986learning}
is the most successful algorithm in AI and machine learning for training deep neural networks. Recently, however, limitations of BP have brought attention back to neuroscience-inspired learning. In particular, it is still unknown whether neural architectures trained via BP will be able to reach a level of intelligence, cognitive flexibility,
and energy consumption comparable to the human brain. 
%, and whether it will be possible to reduce the energy consumption of training a neural network by orders of magnitude. %It is in fact known that training large models requires an amount of energy and computing power that is inaccessible to most business and research groups \citep{Strubell19}. 
This could be solved using alternative learning methods that rely on locally available information, as learning in the brain does. An algorithm with extremely promising properties is \emph{predictive coding (PC)}, an error driven learning algorithm with local updates.

%Because of multiple reasons that will be highlighted in this review, an  algorithm with promising properties is predictive coding (PC), an error-driven learning method with local updates.

PC \citep{rao1999predictive,friston2005theory,srinivasan1982} has emerged as an influential theory in computational neuroscience,  which has a significant mathematical foundation as variational inference, linking it closely with normative theories of the Bayesian brain \citep{knill2004bayesian}, and which provides a single mechanism that can explain many varied perceptual and neurophysiological effects \citep{hohwy2008predictive,auksztulewicz2016repetition,lotter2016deep}, while also postulating a biologically plausible neural dynamics and synaptic update rules \citep{friston2003learning,Lillicrap20,millidge2020relaxing}. 

%such as bistable perception \citep{hohwy2008predictive}, illusory motions \citep{lotter2016deep,watanabe2018illusory}, end-stopping and extra-classical receptive fields in V1  \citep{rao1999predictive}, repetition suppression \citep{auksztulewicz2016repetition}, and top-down attentional modulation of neural activity \citep{feldman2010attention,kanai2015cerebral}, while also postulating biologically plausible neural dynamics and synaptic update rules \citep{friston2003learning,Lillicrap20,millidge2020relaxing}, and detailed implementations at the level of neural microcircuits \citep{bastos2012canonical,shipp2016neural}. 

The fundamental idea of PC is to treat the cortex as performing simultaneous inference and learning on a hierarchical probabilistic generative model, which is trained in an unsupervised setting to predict incoming sensory signals \citep{rao1999predictive,friston2005theory,clark2015surfing}. In such an architecture, at each layer of the hierarchy, top-down predictions emanating from higher layers are matched with and cancel out incoming sensory data or prediction errors from lower layers. Unexplained aspects of the sensory data, in the form of prediction errors, are then transmitted upwards for higher layers of the hierarchy to explain. The transmission of only error information possesses a solid basis in information theory, where it is a way to maximize information transmission per bit given a known model of an information source \citep{spratling2017review,bradbury2000linear}, an important consideration for the brain, heavily optimized by evolution to satisfy tight constraints on energy usage and wire lengths, which thus must transmit and process as little information as possible \mbox{\citep{barlow2001redundancy}.} 

Historically, PC was first proposed for the retina \citep{srinivasan1982}, where neural circuits already subtract away much of the redundant information in the visual stimulus. The same principle was then applied as a general model for cortical processing by \citet{rao1999predictive}, who showed that the model could replicate several well-known responses of neurons in the early visual cortex. The mathematical interpretation of the algorithm as performing variational inference was presented by \citet{friston2003learning} and \citet{friston2005theory}.

PC is also closely related to the more general \emph{free-energy principle} in theoretical neuroscience \citep{friston2006free}, which states that the fundamental drive of the brain is to minimize the variational free energy via both perception (inference and learning) and action. PC networks (PCNs) can be derived as a special case (a ``process theory'') of the free-en\-er\-gy principle assuming a Gaussian generative model and performing inference and learning. The application of PC to robotics and its relationships to classical control theory depend on the third interpretation of the free-energy principle, where free energy is minimized via action, closely linked with the ideas of active inference \citep{friston2017graphical,friston2017active}.
For further reviews of PC, its mathematical foundation, and applications in neuroscience, see \citep{bogacz2017tutorial,buckley2017free,millidge2021predictive}.

Although originating in neuroscience,
%inspired by the success of modern deep learning techniques at learning to perform large-scale and extremely challenging perceptual classification techniques on unstructured data inputs such as images \citep{Krizhevsky2012}, audio \citep{van2016wavenet}, and natural language \citep{Vaswani17,brown2020language},
 a body of literature has investigated how PC can be related and applied to the existing deep learning literature. In this survey, we review this literature, which has developed in the last few years, focusing first on the recently uncovered relationships between the parameter updates in PCNs and in BP-trained artificial neural networks (ANNs), and second on the performance and superior flexibility of PCNs on large-scale deep learning tasks. This superior flexibility, combined with using only local computations ultimately enables a much greater parallelizability of PCNs compared to ANNs, especially on neuromorphic hardware. This greater scalability means that, as ANNs continue to scale \citep{kaplan2020scaling}, the limited memory bandwidth afforded by current GPUs may become increasingly a limiting factor in training, and the greater parallelizability and memory bandwidth of neuromorphic hardware, where computations and memories
 are colocated, may lead to the adoption of PCN-like architectures, which can be efficiently trained on such hardware, leading ultimately to a future of PCNs trained without~BP. 
 
The rest of this survey is organized as follows. In Section 2, we present a general overview of PCNs, first describing their mathematical structure and their training and testing dynamics, and secondly their interpretation as variational inference algorithms. In Section~3, we review the recently uncovered relationships between PC and BP. In Section 4, we discuss the capabilities of PCNs on classification, generation, and reconstruction tasks. In Section 5, we demonstrate that PCNs can function as associative memory models, and in Section~6, we generalize PCNs to arbitrary graph topologies. In Section~7, we review applications of PCNs to problems in control and robotics, and in 
Section~8, we summarize, and we discuss open research challenges for PC in machine learning.

\begin{figure*}[t]
    \centering
    \includegraphics[width=\linewidth]{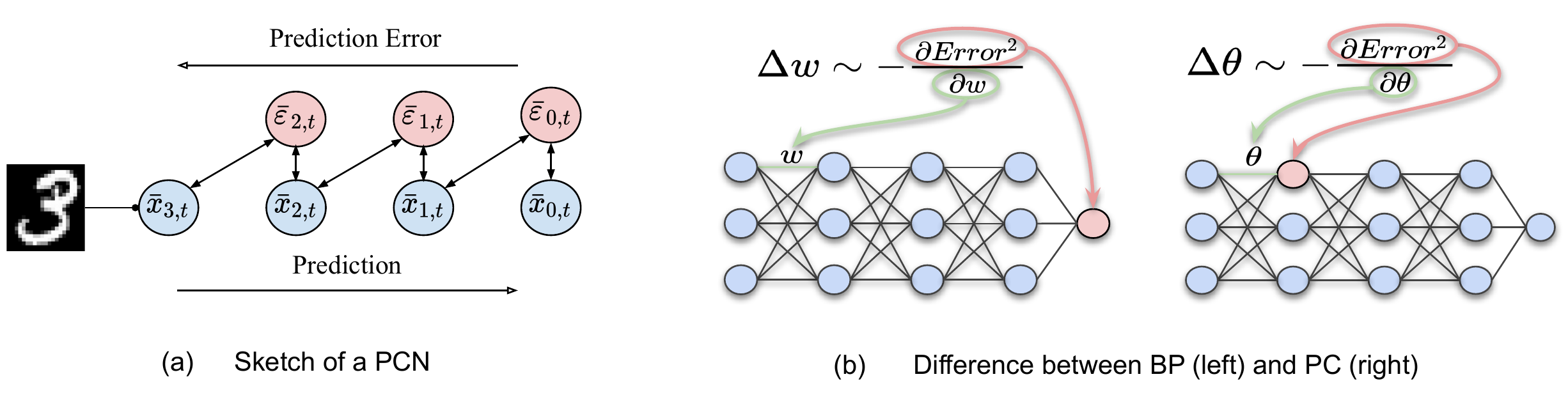}
    \caption{(a): A multilayer PCN trained on a data point of MNIST. Here, the neural activities of a specific layer predict the ones of the previous layer in a forward direction. The error in this prediction is then propagated back down the hierarchy. (b) Difference between the update rules of BP (left) and PC (right). Particularly, the loss function of BP defines an error only on the output layer, and this error is minimized via gradient descent. In very deep networks, this causes weights to be updated to minimize an error that could be dozens of layers away. By contract PC minimizes a local energy function for each layer.}
    \label{fig:1}
\end{figure*}

\section{Overview of Predictive Coding}

We now recall the key concepts of PC. We give an overview of PCNs, which are the PC equivalent of ANNs, and we review the interpretation of PC as variational inference, which provides additional insights into the computations and learning of PCNs. 

\subsection{Predictive Coding Networks}
The mathematical formulation of PC can be interpreted as postulating two kinds of ``neurons''.  The first encodes time-dependent neural predictions and is denoted by $\bar x_t$, while the second encodes prediction errors and is denoted by $\bar \varepsilon_t$. Both quantities are $N$-dimensional vectors that change over time steps $t$ during inference and learning. 
Hierarchical PCNs can be expressed as ANNs, so that PCNs  can be directly compared to ANNs. As such, the value and error nodes are partitioned into $L+1$ layers. We denote by $\bar x_{L,t}$ the neurons of the input layer, and
by $\bar x_{l,t}$ and $\bar \varepsilon_{l,t}$ the neurons of the other layers $l\in\{0,\ldots,L-1\}$.  In this case, a specific stimulus is propagated up the hierarchy, and the goal of every layer of neurons is to predict the value of the following layer according to 
\begin{equation}
\bar \mu_{l,t} = \bar \theta_{l+1} f(\bar x_{l+1,t}),
\label{eq:pcn-forward}
\end{equation}
where $f$ is a non-linearity, and $\bar \theta_{l+1}$ is the matrix of weights connecting layer $l+1$ to layer $l$. The error of this prediction is the difference between the neural activity of a layer and its prediction, i.e., $\bar \varepsilon_{l,t} = \bar x_{l,t} - \bar \mu_{l,t}$, which is then propagated down the hierarchy, and used in the learning process to update the weights of the network. Ultimately, the learning algorithm optimizes a global energy function, defined as the sum of squared prediction errors at each layer:
\begin{equation}
\mathcal{F}_{t} = \frac 1 2 \sum\nolimits_l \| \bar \varepsilon_{l,t} \|^2.
\label{eq:pcn-f}
\end{equation}

\vspace{1ex}
\noindent
\textbf{Training:} During  training, the highest layer is fixed to an input data point $\bar s_{in}$, i.e.,  $\bar x_{L,t} = \bar s_{in}$ for every $t$, and the lowest layer is fixed to a label or target vector $\bar s_{out}$ in the same way. During a process called \emph{inference}, the weight parameters are fixed, and the neural activities are continuously updated to minimize the energy function of Eq.~\eqref{eq:pcn-f} by running gradient descent until convergence, at which point a single weight update is performed. During the weight update, the value nodes are fixed, and the weight parameters are updated via gradient descent on the same energy function. When defining inference and weight update this way, every computation only needs local information to be updated. For a detailed derivation of these equations, refer to \citet{whittington2017approximation}.

\vspace{1ex}
\noindent
\textbf{Testing:} Here, only the highest layer is fixed to the data, so the network infers the label given a test point. This process is equivalent to the inference phase described above: the weight parameters are fixed, and the neural activities are updated until convergence by running gradient descent on the energy function. Note that different works follow different paradigms for interleaving the updates of the neural activities $\bar x_t$ and weights $\bar \theta$. In most works, neural activities are all simultaneously updated for $T$ iterations with the aim of reaching convergence of the inference process, and the weights are updated once upon convergence \citep{whittington2017approximation,salvatori2021associative,millidge2020predictive}. However, in certain cases, it has been noted that updating the weights and activities of different layers in different moments yields a better performance \citep{ororbia20,han2018deep}.

%Usually, the prediction error at the final output layer is equivalent to the loss function used by the equivalent ANN trained with BP. Importantly, the PCN energy function is composed as a sum of layerwise energies, which means that updates for each layer only require local information. In PCNs, the energy is optimized by gradient descent by updating both activations $\bar x_l$ and weights $\bar \theta_l$. The update of the activiations is often called \emph{inference}, while the update of the weights is called \emph{learning}. 

\subsection{Predictive Coding as Variational Inference}

Mathematically, PCNs can also be expressed as va\-ria\-tional inference on hierarchical Gaussian generative models. 
%Although PC can be applied to arbitrary graphs \citep{millidge2020predictive,Salvatori2021}, we here assume a lay\-er-wi\-se structure that mimics the multi-layer perceptrons (MLPs) widely used in deep  learning. 
We define a hierarchy of layers indexed by $l$, where the distribution of activations at each layer is a Gaussian distribution with a mean given by a nonlinear function of the layer below 
 $f(\bar x_{l+1})$ with parameters $\bar \theta_{l+1}$ and an identity covariance $I$:
\begin{align*}
    \label{pc_gen_model}
    p(\bar x_{0,t} \dots \bar x_{L,t}) &= p(\bar x_{L,t}) \prod\nolimits_{l=0}^{L-1} p(\bar x_{l,t} | \bar x_{l+1,t}) \\
    p(\bar x_{l,t} | \bar x_{l+1,t}) &= \mathcal{N}(\bar x_{l,t} ; \bar \theta_{l+1} f(\bar x_{l+1,t}), I)\,.
\end{align*}
The PCN can then be ``queried'' by conditioning on a data or label item. For instance, suppose that the input layer $l=L$ is fixed to some data item $\bar s_{in}$. We then wish to infer the state of the rest of the network given this conditioning $p(\bar x_{0,t} \dots \bar x_{L-1,t} | \bar x_L)$. This inference problem can be solved by variational inference \citep{beal2003variational,wainwright2008graphical,jordan1998introduction}. In general, variational inference approximates an intractable inference through an optimization problem, where the parameters of an approximate \emph{variational posterior} distribution $q$ are optimized, so as to minimize its distance from the optimal posterior $p$. This optimization procedure is performed by minimizing an upper bound on this divergence known as the \emph{variational free energy} $\mathcal{F}_t$.

%\begin{align}
%    q^* &= %\underset{q}
%    {\text{argmin}_q} \, \, \mathcal{F} \\
%    \mathcal{F} &= KL \big[q(\bar x_1 \dots %\bar x_L | \bar x_0) || p(\bar x_0 \dots \bar x_L) \big] \\ &\geq KL[q(\bar x_1 \dots \bar x_L | \bar x_0) || p(\bar x_1 \dots \bar x_L | \bar x_0)]
%\end{align}
In PCNs, we assume that the variational posterior is factorized into independent posteriors for each layer $q(\bar x_{0,t} \dots$ $\bar x_{L-1,t}) = \prod_{l=0}^{L-1} q(\bar x_{l,t})$ and, combined with the Laplace approximation \citep{friston2007variational}, this allows us to considerably simplify the expression for the free energy into a sum of squared prediction errors (see \citet{buckley2017free}), equivalent to the one in Eq.~\ref{eq:pcn-f} up to an additive constant:
\begin{align}
    \mathcal{F}_t &\approx \sum\nolimits_{l=0}^{L-1} \log p(\bar x_{l,t} | \bar x_{l+1,t}) 
    \approx \sum\nolimits_{l=0}^{L-1}  \| \bar \varepsilon_{l,t}\|^2\,,
\end{align}
where $\bar \varepsilon_{l,t} = \bar x_{l,t} - {\bar \theta_{l+1}} f(\bar x_{l+1,t})$ is the ``prediction error'' for each layer. When applied to ANNs, we typically assume that the layerwise dependencies are parametrized by a parameter matrix $\bar \theta_{l+1}$, which corresponds to the weights in an ANN. Then, both the activations $\bar x_{l,t}$ and weights can be updated as a gradient descent on the free energy,
\begin{align}
    {d \bar x_l}/{dt} &\propto - {\partial \mathcal{F}_t}/{\partial \bar x_{l,t}} \\
    {d \bar \theta_l}/{dt} &\propto - {\partial \mathcal{F}_t}/ {\partial \bar \theta_l}\big|_{\bar x = \bar x^*_{l,t}}\,.
\end{align}

PCNs operate in two phases, where first the activation means $\bar x_{l,t}$ are updated to minimize the free energy until they reach an equilibrium, and then the weights $\bar \theta_l$ are updated for a single step given the equilibrium values of the activations $\bar x^*_{l,t}$. These phases are known as \emph{inference} and \emph{learning}.

In the context of ANNs, PC recasts the feedforward pass in an ANN as an inference problem over the activations of the layers of an ANN given some conditioning on either input or output layers, or both, where the uncertainty about the ``correct'' activations is assumed to be Gaussian around a mean given by the top-down prediction from the layer above. Importantly, this inference problem is solved dynamically during each inference phase, and the conditioning variables can be varied flexibly depending on the desired task. This enables the PCN to use its learned generative model (encoded in the weights $\bar \theta_l$) to be repurposed for different inference problems at run-time, and accounts for the superior flexibility of PCNs over ANNs demonstrated in recent work.

\section{Predictive Coding and Backpropagation}

%Historically, different algorithms have been proposed to train neural networks, with the goal of overcoming the few limitations of BP \citep{rbm,hebb49,Hinton06,dbm}. Exploring the similarities between neuroscience-inspired algorithms and BP has in fact been an active area of recent research \citep{Lillicrap20}, as the desire of finding a novel learning algorithms that aims to replace BP in some tasks cannot be blind of the impressive performance reached by BP on pattern recognition applications.

%Although the backpropagation (BP) of error \citep{rumelhart1986learning} has achieved fantastic success at training large-scale deep neural networks to perform a huge variety of tasks \citep{Krizhevsky2012,Vaswani17,lstm,he2016deep}, from the perspective of neuroscience, it has often been criticized for its biological implausibility, due to its use of non-local information to compute gradients and its requirement to separate forward and backward phases \citep{Lillicrap20,crick1989recent,whittington2019theories}. Although how the brain performs credit assignment, the rule governing information passing and synaptic updates, in a local fashion remains unknown, proposing biologically-plausible and neuroscience-inspired algorithms for credit assignment to replace BP has been an active area of recent research \citep{xie2003equivalence,lillicrap2016random,lee2015difference,liao2016important,scellier17,scellier2018generalization,millidge2020relaxing,millidge2020activation,akrout2019using,whittington2019theories}.

%\smallskip 
%\noindent
%\textbf{Relationship between PC and BP.}
Recently, multiple results have explored similarities and relationships between PC and BP, showing that PC can closely approximate or exactly perform BP under certain conditions on supervised learning tasks. Firstly, it has been shown that PC well approximates the parameter update of BP on multi-layer perceptrons (MLPs), albeit under some strict conditions \citep{whittington2017approximation}. This result has been recently extended in two orthogonal directions: it has been shown that PC converges to BP not only on MLPs, but also on any computational graph \citep{millidge2020predictive}. For these results to hold, one of two conditions must be met: either the activity values remain very close to their feedforward pass values such that the prediction error is small, or else the layerwise derivatives must be held fixed to their feedforward pass values and the network run to equilibrium. Moreover, experimental results also empirically show that PC approximates BP updates under less restrictive conditions, i.e., a small output error is enough, and the energy does not have to be completely converged.  An exactness result  also holds, as a variation of PC, called Z-IL, performs exact BP on MLPs if the weights are updated after the first non-zero inference step at each layer when the network activations are initialized to their feedforward pass values \citep{Song2020}. These results also extend to Z-IL being able to exactly replicate the parameter update of BP on any computational graph \citep{salvatori2021any}. The advantage of these exactness results are twofold: first, for Z-IL, they require only a small number of time steps to perform a full update of the parameters, and empirical results show that Z-IL is almost as efficient as BP, at the cost of requiring complex control logic to synchronize parameter updates to occur at the correct times across layers. Second, it provides a novel (but equivalent) implementation of BP, which is able to learn via local computations. All  these results are experimentally validated on multiple architectures, such as LSTMs \citep{lstm}, transformers \citep{Vaswani17}, and ResNets \citep{he2016deep}. A historical sketch of these results is given in Fig.~\ref{fig:gen}(a).

\section{Performance of Predictive Coding} 

In this section, we review recent results obtained by PCNs on classical computer vision benchmarks. Particularly, we briefly review results in image classification and generation.

\vspace{1ex}
\noindent
\textbf{Classification:} The connection to BP for supervised learning suggests that PCNs should perform well on image classification tasks. This is indeed the case: the first formulation of PC for supervised learning, equivalent to the one described in the preliminary section, shows that PC is able to obtain a performance comparable to BP on small multilayer networks trained on MNIST \citep{whittington2017approximation}. A similar result was also obtained using a variation of PC not restricted to be trained using mean squared error as an energy function. It is in fact possible to define a generalization of IL, which uses layer-specific loss functions  \citep{ororbia2019biologically}. This variation, called  \emph{local representation alignment}, is able to reach a competitive performance with BP on MNIST and FashionMNIST. On more challenging tasks, a deep convolutional PCN is able to achieve a performance similar to BP on complex datasets, such as CIFAR10, and acceptable results on ImageNet \citep{han2018deep}. This model updates the value nodes of one layer at a  time, multiple times via lateral recurrent connections, before performing a weight update. It is intuitively similar to a deep convolutional network trained using the Z-IL algorithm, augmented with lateral connections.

\begin{figure*}[t]
    \centering
    \includegraphics[width=\linewidth]{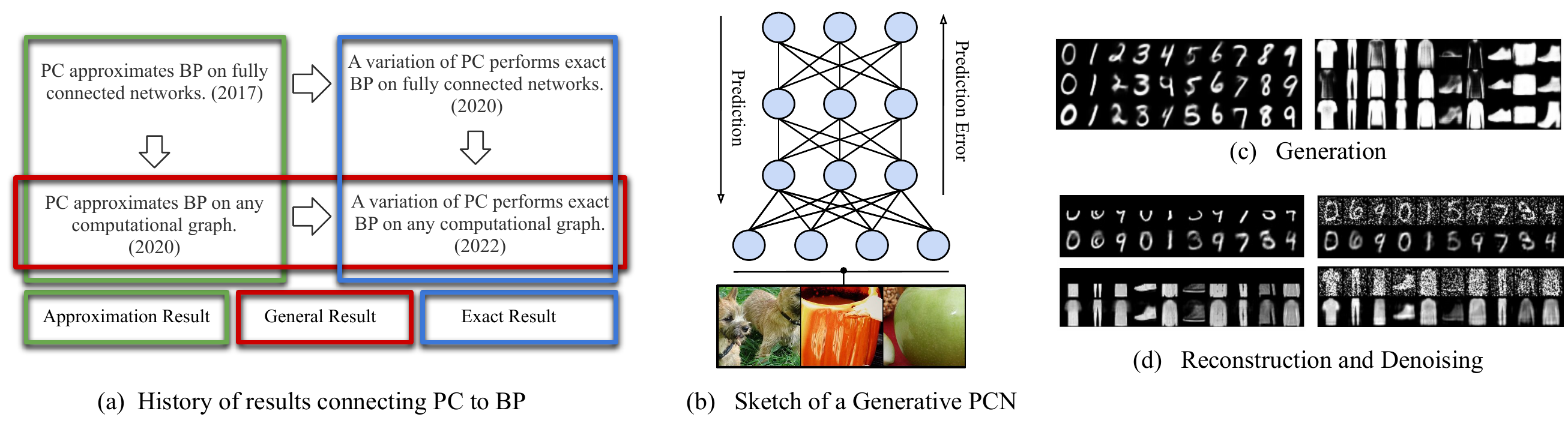}
    \caption{(a) Historical and conceptual sketch of the results unifying predictive coding (PC) and backpropagation (BP). (b) Sketch of a generative PCN. In contrast to networks trained for classification, the input image is presented in the first layer of the network. The energy minimization updates the weights to get zero (or low) error on it. (c) Examples of generated MNIST and FashionMNIST using a generative PCN. (d)  Examples of reconstructed (left) and denoised (right) MNIST and FashionMNIST images using a generative PCN.  Figures (c,d) are taken from the original papers, i.e.,  \protect\citep{ororbia20} and  \protect\citep{salvatori2022learning}, respectively.}
    \label{fig:gen}  
\end{figure*}

\vspace{1ex}
\noindent
\textbf{Generation:} Above, we have reviewed the connection between PC and BP for image classification, which implies that, with some effort, it should be possible to use PC to train classifiers on large-scale datasets. PC is, however, a generative model, as suggested by its formulation as variational inference \citep{rao1999predictive,friston2005theory}. This implies that the PC models surveyed in the previous section can also be used for data generation from labels, as long as certain regularizations are applied due to the ill-posed nature of the inverse problem \citep{sun2020predictive}. Additionally, PCNs can also be used directly as generative models due to their interpretation as probabilistic graphical models by swapping the ``direction'' of the network, so that the label is treated as the ``input'', and the data are treated as the ``output''.
The basic architecture used to perform generative tasks resembles the decoder part of autoencoders, and is sketched in Fig.~\ref{fig:gen}(b).
More complex models, which have a similar structure but are augmented with different kinds of connections, have been shown to generalize to unseen images \citep{ororbia20,salvatori2022learning}. Particularly, \citeauthor{ororbia20} \citeauthor{ororbia20} present three generative models: the first one is a novel model with recurrent connections, while the second and the third are implementations of Rao and Ballard's original PC \citep{rao1999predictive}, and of a model designed by \citeauthor{friston2008hierarchical} \citep{friston2008hierarchical}. The extensive experiments show that generative PCNs are able to successfully generate novel black and white images of different datasets, as shown in Fig.~\ref{fig:gen}(c). A qualitative evaluation against standard baselines in machine learning shows that this method obtains comparable results. Hence, an interesting future direction is to directly test the generation capabilities of large-scale PCN equivalents to deep convolutional networks on challenging image datasets such as ImageNet.

%Hence, an interesting future direction is to use more complex kinds of layers, such as convolutions, to test their generation capabilities on challenging datasets of colored images.

An important quality of generative PCNs is their ability to generalize well on novel tasks. This suggests that they learn an internal probabilistic representation of the dataset, and can apply it when tested on tasks that they were not trained for. This differs from ANNs, which exhibit a generalization across data from the same task but fail to generalize \emph{across tasks}. The generalization capability of PCNs is more equivalent to meta learning. The greater flexibility of PCNs originates from their inference phase. For instance, it is possible to provide a PCN with half a test image and let the network infer the missing pixels via running energy minimization, or to present a corrupted data point (e.g., with Gaussian noise), and ask the model to clear it, as shown in Fig.~\ref{fig:gen}(d). 
%or to present the network with a partial ambiguous image and a label, and use the label information to fill in the remaining image using the contextual label information. 
Importantly, PCNs can accomplish these task even if not directly trained to do so, unlike ANNs, which must be trained to perform each specific task \citep{ororbia20,salvatori2021associative,salvatori2022learning}.

\section{Associative Memories}

\begin{figure*}[t]
    \centering
    \includegraphics[width=\linewidth]{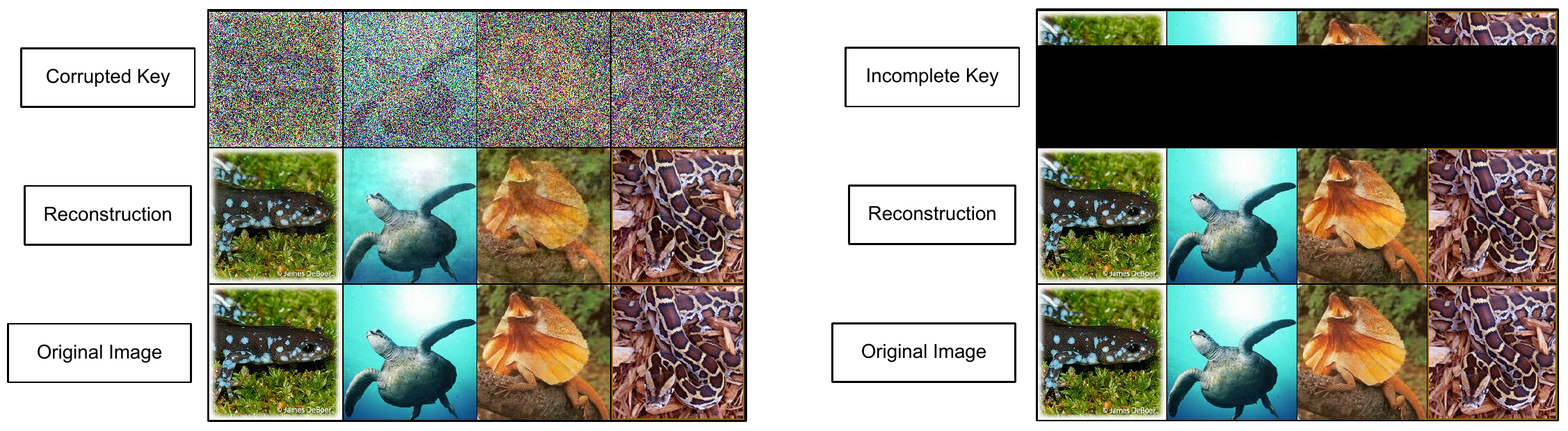}
    \caption{Examples of retrieved ImageNet pictures when presenting a corrupted key with gaussian noise of mean zero and variance $\eta = 2.0$ (left) and an incomplete key, where only $\frac 1 4$ of the original pixels were provided. Particularly, $100$ images were stored in this example.}
    \label{fig:am}  
\end{figure*}

It has also been recently demonstrated that generative PCNs also function as associative memories. In machine learning, the task of an associative memory model is to store and retrieve data points. When presented with a corrupted or incomplete variation of a stored data point, a good associative memory model has to detect the uncorrupted memory and return it as an output. Generative PCNs are able to store and retrieve complex memories, such as ImageNet pictures \citep{salvatori2021associative}. This is done by training the model on a subset of  ImageNet, and retrieving the original data points via energy minimization when providing highly corrupted or incomplete variants of them. While the fully connected PCN structure does not allow generalization to unseen data points on complex images such as ImageNet, their retrieval capacity demonstrates their ability to generate high-quality reconstructions, see Fig.~\ref{fig:am} for examples.
This model has been extensively compared with different associative memory models, such as continuous-state Hopfield networks \citep{ramsauer21} and autoencoders trained with BP \citep{radhakrishnan19}. In both cases, the PC based memory model has outperformed its classic counterparts, showing a retrieval robustness superior to any other baseline. This model also possesses a high degree of biological plausibility, as it has been hypothesised that the brain stores and retrieves memories using a PC architecture where the hippocampus sends fictive prediction errors to the sensory neurons via hierarchical networks in the neocortex, which are minimized by memory retrieval \citep{Barron20}. 

\section{Learning on Arbitrary Graph Topologies}

Learning on networks of any structure is not possible using BP, where information first flows in one direction via the feedforward pass, and the error in the reverse direction during the backwards pass. Hence, a cycle in the computational graph of an ANN trained with BP would cause an infinite loop. While the problem of training on some specific cyclic structures has been partially addressed using BP through time \citep{lstm,rumelhart1986learning,williams1989learning}  on sequential data, the restriction to hierarchical architectures may present a limitation to reaching brain-like intelligence, since the human brain has an extremely complex and entangled neural structure that is heterarchically organized with small-world connections \citep{avena18}---a topology that is likely highly optimized by evolution. 
%This shape of structural brain networks, shown in Fig.~\ref{fig:arbitrary}, generates a unique communication dynamics that is fundamental for information processing in the brain, as different aspects of network topology imply different communication mechanisms, and hence perform different tasks \citep{avena18}. 
Hence, a recent direction of research aimed to extend learning to arbitrary graph topologies. A popular example is the \emph{assembly calculus} \citep{Papadimitriou20}, a Hebbian learning method that can perform different operations implicated in cognitive phenomena. However, Hebbian learning methods cannot perform well compared to error-driven ones such as BP \citep{movellan1991contrastive}. PC, however, has both the desired properties that allow high-quality representation learning on arbitrary graph topologies: it is error-driven, and only learns via local computations. Moreover, it has been shown that it is possible to perform generation and classification tasks on extremely entangled networks, which closely resemble brain regions \citep{salvatori2022learning}. This enables a more general learning framework, which converges to a global solution via energy minimization that can perform multiple tasks simultaneously, such as classification and generation, but also to develop novel architectures, optimized for a single specific task. Tested on generation, reconstruction, and denoising tasks, this model has been shown to have a performance superior or comparable to standard autoencoders.

\section{Predictive Coding for Control and Robotics}

\begin{figure}[t]
    \centering
    \includegraphics[width=\columnwidth]{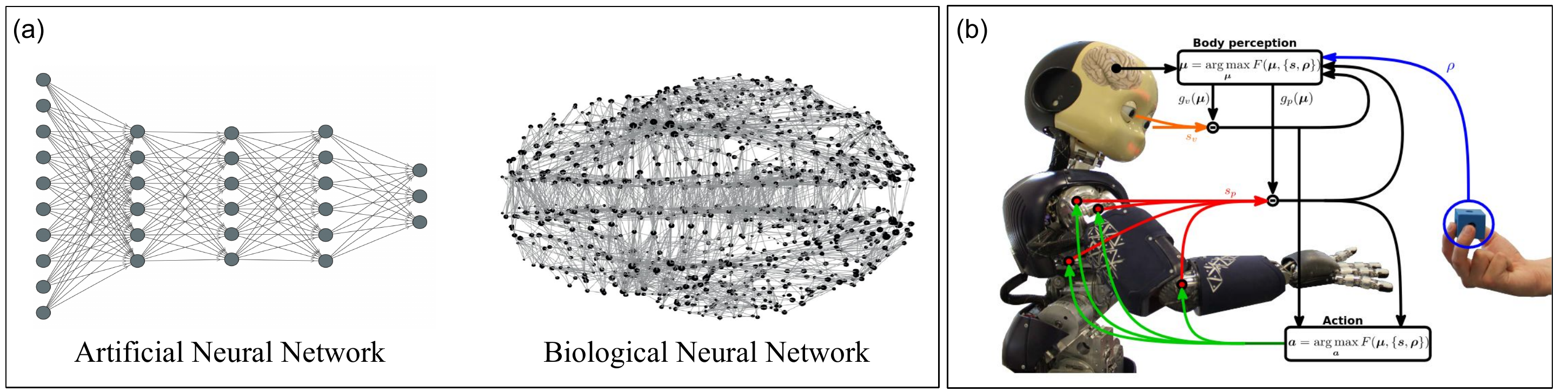}
    \caption{ (a)
    Difference in topology between an ANN (left) and a sketch of a network of structural connections that link distinct neuronal  elements in a brain (right). Figure taken from \protect\citep{salvatori2022learning}. (b) Graphical representation of the joint minimization of free energy by action and control to enable a simultaneous state estimation and action selection in a robotic grasping task with a humanoid robot. Figure taken from \protect\citep{oliver2021empirical}.
    }
    \label{fig:arbitrary}
\end{figure}

In line with the free-energy principle in considering both perception and action as emerging from an imperative to minimize free energy, PC methods can also be directly applied to control problems, with close links to classical control theory, and have been applied productively to problems in robotics. Although the free energy does not explicitly include an action term, it includes one implicitly through the dependence on data $\bar x_0$ which could depend on actions $a$. This dependency can be computed explicitly by using the chain rule \citep{friston2009reinforcement,friston2011optimal},
\begin{align}
    {da}/{dt} \propto -{\partial \mathcal{F}}/{\partial a} = - {\partial \mathcal{F}}/{\partial \bar x_0} \cdot{\partial \bar x_0}/{\partial a}\,,
\end{align}
where ${\partial \bar x_0}/{\partial a}$ is known as a \emph{forward model}, which explicitly quantifies how data depend on observations. In the case of linear generative models and using generalized coordinates of motion \citep{baltieri2019pid}, PC has been shown to be equivalent to Proportional-Integral-Derivative (PID) control, a widely used and effective method in classical control theory \citep{johnson2005pid}. PC also provides a generalization of PID for both additional dynamical orders (i.e., using fourth- and higher-order derivatives) as well as instant generalizations to non-identity and ultimately nonlinear dynamics.

PC methods have also been widely used in robotics. The generative model in PC can be set to model the dynamics of a system and then torques can be inferred to realize a desired motion \citep{lanillos2018adaptive}. PC has thus been applied to a variety of robotics problems \citep{lanillos2018adaptive} as well as drone and quadcopter control \citep{meera2020free,meera2021brain}. An additional advantage of PC is that it can also be used for state estimation \citep{pezzato2020novel,oliver2021empirical} (including with high dimensional image inputs \citep{sancaktar2020end}, providing a joint solution of state estimation and control, as depicted in Figure \ref{fig:arbitrary}(b). For a recent full review of this literature, see \citet{lanillos2021active}.

More generally, if we consider PCNs as embodying an implicit probabilistic graphical model, and interpret some nodes of the PCN as ``action nodes'', then given that other nodes can be fixed to a set of desired outcomes of control, then the PC inference algorithm will infer the actions consistent with the conditioned outcomes. Explorations of this idea are presented in \citep{bogacz2020dopamine,kinghorn2021habitual}, and this mechanism is simply an example of the active inference \citep{friston2017graphical,millidge2020relationship} and control as inference \citep{attias2003planning,toussaint2006probabilistic,levine2018reinforcement} frameworks, which interpret control problems as probabilistic inference problems, which simply require inferring the correct action or action sequences conditioned on achieving a high reward or desired state trajectory.

\section{Summary and Open Challenges}

Both the theory and practice of PC have advanced substantially over the last few years, with both an important set of theoretical results revealing a close connection with BP being developed, as well as significant empirical strides being made in the capacity of PCNs to perform successfully on large-scale machine learning benchmarks, often with a performance comparable to equivalent ANNs. 
%Given the close connections of PCNs with neuroscientific theories of cortical function, these are significant advances since it is essentially the first time a neuroscientifically derived and more biologically plausible theory has demonstrated the ability to scale to the large-scale tasks currently handled by ANNs and which are also faced by the brain.

While current research has shown that PC is closely connected to BP and that PCNs can often match the performance of BP-trained models on a variety of machine learning benchmarks, there are only a few cases (such as associative memories) where PCNs perform demonstrably \emph{better} than comparable ANNs. Thus, at present, PC is not yet used in industrial applications. However, the promising properties highlighted in this survey strongly suggest that this will change in the next years. Hence, future research should be directed at finding applications where the unique properties of PC can be utilized to outperform existing methods, as well as theoretical research exploring these properties further.

One unique property of PC, which is not shared by BP-trained ANNs, is its superior flexibility due both to its inference phase and mathematical interpretation as a generative model, and to the local computations that allow training on any graph structure. The first means that PCNs can flexibly perform different inference tasks given the same network (while an ANN would have to be trained separately for each task), and also allows PCNs to natively handle missing data, while ANNs need heuristic imputation schemes. The second, could allow new progress in neural architecture search. Beyond this, it is worth noting that essentially all current benchmarks, layers, initializations, and other ``tricks'', have been heavily optimized specifically for ANNs, while no such optimization work has been performed for PCNs, thus indicating that an important line of future research will be to devise similar PCN-specific ``tricks'' to improve performance. This line of research is of vital importance to scale the applicability of PCNs.

An additional unique property of PCNs compared to ANNs is the locality of their weight updates and hence greater parallelizability compared to ANNs. If exploited properly, this may enable a substantially more efficient training of extremely large-scale PCN models by reducing the communication and wait time requirements induced by the sequential backward step of BP, as well as lend itself to an efficient implementation on neuromorphic hardware.

Finally, while theoretical results indicate close connections with BP, these only apply under certain conditions. Future work may investigate the behavior of PCNs when these conditions are relaxed, and whether it has any advantages or different properties compared to standard BP.

\section*{Acknowledgments}
This work was supported by the Alan Turing Institute under the EPSRC grant EP/N510129/1, 
by the AXA Research Fund, the EPSRC grant EP/R013667/1, and by the EU TAILOR grant. We also acknowledge the use of the EPSRC-funded Tier 2 facility
JADE (EP/P020275/1) and GPU computing support by Scan Computers International Ltd.

%\bibliographystyle{ieeetr}
%\bibliography{apacite}
\bibliography{references}

\end{document}